\documentclass[lettersize,journal]{IEEEtran}
\usepackage{amsmath,amsfonts}
\usepackage{algorithmic}
\usepackage{algorithm}
\usepackage{array}
\usepackage[caption=false,font=normalsize,labelfont=sf,textfont=sf]{subfig}
\usepackage{textcomp}
\usepackage{stfloats}
\usepackage{url}
\usepackage{verbatim}
\usepackage{graphicx}
\usepackage{cite}
\usepackage{float}
\usepackage{caption}
\usepackage{booktabs}

\begin{document}

\title{Least Squares Maximum and Weighted Generalization-Memorization Machines}

\author{Shuai Wang, Zhen Wang and Yuan-Hai Shao$^*$
\thanks{S. Wang is with the School of Mathematics and Statistics, Hainan University , Haikou, China.
(e-mail: wangshuai282615@163.com).}
\thanks{Z. Wang is with the School of Mathematical Sciences, Inner Mongolia
University, Hohhot, 010021, P.R. China. (e-mail: wangzhen@imu.edu.cn).}
\thanks{Y.H. Shao (*Corresponding author) is with the Management School, Hainan University, Haikou, P.R. China. (e-mail:shaoyuanhai21@163.com).}
\thanks{Manuscript received xx, xx; revised xx, xx.}
}

\maketitle

\begin{abstract}
  In this paper, we propose a new way of remembering by introducing a memory influence mechanism for the least squares support vector machine (LSSVM). Without changing the equation constraints of the original LSSVM, this mechanism, allows an accurate partitioning of the training set without overfitting. The maximum memory impact model (MIMM) and the weighted impact memory model (WIMM) are then proposed. It is demonstrated that these models can be degraded to the LSSVM. Furthermore, we propose some different memory impact functions for the MIMM and WIMM. The experimental results show that that our MIMM and WIMM have better generalization performance compared to the LSSVM and significant advantage in time cost compared to other memory models.
\end{abstract}


\section{Introduction}
\IEEEPARstart{Z}{ero} experience risk, also known as memory of training data, has been widely researched and discussed in machine learning \cite{memorization,10.1145/3357713.3384290,VAPNIK2021108018}.
Traditional learning machines require to classify the training samples correctly as much as possible, but it is prone to fall into the overfitting problem.
Therefore, to avoid overfitting, we commonly use regularization techniques but also reduce the memory ability, e.g.support vector machines (SVMs) \cite{1995Support}. However, more powerful tools have been proposed in machine learning based on the zero empirical risks. For instance, 
Deep Neural Network (DNN) \cite{memorization,2017A,cohen2019dnn} has a structure of multiple hidden layers. Each neuron receives inputs from the neurons in the previous layer and generates outputs that serve as inputs to the neurons in the next layer. And each hidden layer contains multiple neurons  to achieve the almost zero empirical risk.
It is also realized by Recurrent Neural Network (RNN) \cite{1990Finding,pmlr-v119-yang20j,hinton2015distilling,Goodfellow-et-al-2016} which is a neural network model commonly used in sequential data processing. Compared to traditional feed-forward neural networks, RNN considers temporal dependencies when processing sequential data. Information is allowed to be passed from the current time step to the next time step. This recurrent structure allows RNN to process sequence inputs of arbitrary length and to capture temporal dependencies in the sequence. 
The Long Short-Term Memory (LSTM) \cite{bishop1995neural,1997Long,10.1145/3446776}  is a particular RNN for solving the long-term dependency problem in RNN. Unlike the traditional RNN, the LSTM model introduces three gates (input gate, forget gate and output gate) and a memory unit to effectively capture and remember critical information in long sequences. 
Unlike the LSTM model of memory, Devansh Arpitet al \cite{2017A} investigated the role of memory in deep learning, linking it to ability, generalization, and adversarial robustness. It is also shown that the training data itself plays an important role in determining the degree of memory. 
Zhang et al \cite{Yang2023ResMemLW}.explored a new mechanism to improve model generalization through explicit memory and proposed the residual memory (ResMem) algorithm, a new approach to augment existing prediction models (e.g., neural networks) by fitting the residuals of the model with $k$-nearest-neighbor based moderators. 

Indeed, memory systems have been widely explored by researchers to enhance memorization capabilities in various domains. For instance, in the field of machine learning and artificial intelligence, memory mechanisms have been proposed to assist learners in remembering and revising learning tasks \cite{2020WanShanshan,2020CullyAntoine,clayton2001elements}.Rafferty et al. \cite{PMID:26400190} presented an observable Partially Observable Markov Decision Process (POMDP) planning problem to address memory tasks \cite{8481496,Androulakis2001}, while Settle and Meeder \cite{settles-meeder-2016-trainable} developed a trainable memory retention model that optimizes revision schedules for effective memorization.
In the realm of deep reinforcement learning, researchers have explored novel methods and optimal policies, elevating the efficiency and engagement of learners \cite{9130935,9031418,upadhyay2018deep} .
In other works related to memory, researchers have focused on statistical characteristics of learners' memory behavior rather than just time-series features \cite{Anderson1996ACT,NIPS2009_6bc24fc1}. This approach has been extended to consider forgetting mechanisms and spaced repetition to improve memory retention \cite{10059206, 10.1145/3534678.3539081}. By transforming the optimization problem into a stochastic shortest path problem, these methods aim to enhance the learning process through efficient memory utilization and forgetting strategies \cite{Maddox2011TheRO,DBLP,DBLP:journals/corr/abs-2102-04174}.

Recently, Vapnik and Izmailov \cite{1995Support,2021Reinforced} studied the memory problem of SVMs and introduced two RBF kernels in the SVMs to improve their memory capability, called SVM$^m$. 
The two RBF kernels, one for generalization and one for memory, are used to memorize the training samples
by properly tuning their parameters to achieve zero empirical risk and have a well generalization performance.
Subsequently, a generalization-memorization machine (GMM) \cite{wang2022generalizationmemorization,Smola1998LearningWK} presented a more general model and explained the mechanism of SVM$^m$ more clearly.
 \\ \indent
 In this paper, another new memory mechanism is proposed.
 It contains two memory models by the least squares sense, i.e., a Maximum Impact Memory Model (MIMM) and a Weighted Impact Memory Model(WIMM). 
  Their learning rate are much faster than the GMM and SVM$^m$,while guaranteeing their zero empirical risks.
 \\ \indent
 The main contributions of this paper are as follow:
 \begin{itemize} 
  \item For the memory problem, we proposed the maximum memory impact (MIMM), which uses only the nearest training points for test point judgments and gives a sufficient condition for the empirical risk of the model to be zero.
  \item For the MIMM model, we constructed a memory influence function suitable for the model to ensure the memory capacity of the model.
  \item We provide a clearer interpretation of the memory kernel of the model and derivatively give conditions for the model to degenerate to LSSVM.
  \item Compared with other memory models, the two memory models we proposed, WIMM and MIMM, are shorter in terms of time cost in memorizing the same learning task machines. 
  \end{itemize}
  The next section provides a brief overview of the development of Support Vector Machines (SVM) and Least Squares Support Vector Machines (LSSVM). It also reviews the GMM models.
  The third section introduces the new objective function and the novel memory mechanism. This includes discussing memory cost and impact functions and how they contribute to solving the MIMM and WIMM models.
  The last section presents the numerical experiments conducted to validate the proposed MIMM and WIMM models. Conclusions drawn from these experiments are also discussed in this section

\section{Review}

Consider a binary classification problem in a n-dimensional real space $\mathbb{R}^n$.
The training set is given by T = $\{(\text{x}_i,y_i) | i = 1,2,... ,m\}$,
where $\text{x}_i \in \mathbb{R}^n$ is the $i$th sample, and $y_i \in \{+1,-1\}$ is the corresponding label.
The training samples and their labels are organized into matrix \textbf{X}$\in \mathbb{R}^{n \times m}$
and diagonal matrix \textbf{Y} with diagonal elements $\textbf{Y}_{ii}=y_i$ $(i=1,... ,m)$, respectively.

SVM \cite{1995Support,788640,Vapnik2006EstimationOD} deals with this  binary classification problem by finding a pair of  parallel hyperplanes in the feature space, where the margin is maximized to separate the two classes as much as possible.
Schölkopf et al \cite{2000New}. proposed a new class of regression and classification models based on the SVM, in which a parameter $\nu$ was introduced to not only effectively controls the number of support vectors but also suit for different data distributions well.
Twin Support Vector Machine (TWSVM) was introduced by Jayadeva et al. \cite{Jayadeva2007Twin}. The TWSVM approach aims to identify a pair of non-parallel hyperplanes that can effectively solve the classification problem, resulting in a reduced problem size compared to traditional SVMs.
To further accelerate the learning speed of SVMs,the Least Squares Support Vector Machine (LSSVM) \cite{2002Least,zhang2017understanding} was proposed by J.A.K. Suykens et al. Due to the equation constraints in the LSSVM formulation, it requires to solve a system linear equations rather than the quadratic programming problem in the SVM. However, the zero empirical risks are guaranteed in neither of these SVMs. Recently, Vapnik and Izmailov \cite{1995Support,2021Reinforced,Belkin_2019} proposed a new kernel function consist of two Gaussian kernels as $K(x,x')=\tau \exp \{-\sigma ^2(x-\acute{x})^2\}+(1-\tau) \exp \{-\sigma ^2_*(x-\acute{x})^2\}$ (where $ 0 \leq \tau \leq 1$,and$\sigma_*\gg \sigma $). This kernel function could greatly improve the memory ability of SVM.

To memorize all the training samples,Wang et al. \cite{wang2022generalizationmemorization}  proposed a generalization-memorization machine(GMM) under the principle of large margins, and this mechanism can obtain zero empirical risk easily.
Hard Generalization-Mem\\
-orization Machine (HGMM) \cite{wang2022generalizationmemorization} constructed a classification decision with $f(\text{x}) = <\text{w}, \varphi (\text{x})>+b+\sum\limits_{i = 1}^{m} y_i c_i\delta (\text{x}_i,\text{x})$, and $\text{w} \in \mathbb{R}^d$ and $b \in \mathbb{R}$ by solving
\begin{equation}\label{hgmm}
  \begin{split}
      \min_{\text{w},b,c } \quad 
      &\frac{1}{2}  \|\text{w} \|^2 + \frac{\lambda}{2}\|c\|^2
      \\
      {\rm s.t.} \quad 
      &y_i(<\text{w},\varphi (\text{x}_i)>+b+\sum_{j = 1}^{m} y_j c_j\delta (\text{x}_i,\text{x}_j)) \ge 1,
      \\
      & i =1,...,m,
\end{split}
\end{equation}
where $<\cdot ,\cdot >$ denotes the inner product, $\varphi (\cdot)$ is the mapping, and $\lambda $ is the positive parameter, $c=(c_1,...,c_m)^\top$ denotes the memory cost of the training sample, $\delta (\text{x}_i, \text{x})$ is a memory impact function that we define in advance. For a new sample x, if f(x) > 0, it is classified as positive class with $y = +1$, otherwise it is classified as negative class with $y = -1$. In general, we can solve the pairwise problem of (\ref{hgmm})
\begin{equation}\label{hgmm:qp}
  \begin{split}
    \min_{\alpha} \quad 
    &\frac{1}{2}\alpha ^\top \textbf{Y}(K(\textbf{X},\textbf{X})+\frac{1}{\lambda}\bigtriangleup  \bigtriangleup^\top)\textbf{Y}\alpha-\textbf{1}^\top\alpha ,
    \\
    {\rm s.t.} \quad 
    &\textbf{1}^\top\textbf{Y}\alpha =0,\alpha \ge 0,
\end{split}
\end{equation}
where $\alpha \in \mathbb{R}^m$ is a Lagrangian multiplier vector, $K(\cdot,\cdot) =<\varphi(\cdot), \varphi(\cdot)>$ is a kernel function, and $\textbf{1}$ is a vector with the appropriate dimension. Specifically, a new sample x will be classified as +1 or -1 depending on the decision
\begin{equation}\label{des}
    f(\text{x}) = \sum\limits_{i = 1}^{m} y_i\alpha _i K(\text{x}_i,\text{x})+b+\sum\limits_{i = 1}^{m} y_i c_i\delta (\text{x}_i,\text{x}).
\end{equation}
Furthermore, by finding a non-zero component $\alpha _k$ in the solution $\alpha $(\ref{hgmm:qp}) of the problem, we obtain $b = y_k-y_k\sum\limits_{i = 1}^{m} y_i(\alpha _i \\K(\text{x}_i,\text{x}_k)+ c_i\delta (\text{x}_i,\text{x}_k))$.

The above HGMM has good generalization ability for many problems, but it is time consuming for big data problems and cannot always classify all training samples quickly. For a memory problem, we not only need to be able to remember the training samples quickly, but also need to give labels quickly during testing. The optimization problem (\ref{hgmm}) with a memory cost function is a practical path to memorize the training samples. We consider the case where the constraints of this optimization problem are equivocal and propose a new construction on the optimization problem. From this perspective, for our machine learning model, we can solve the problem by solving a system of linear equations. In other words, we have a faster memory effect compared to HGMM, regardless of the complexity of the corresponding learning task. Also, we consider a new type of memory different from the weighted memory in HGMM and propose several constructions of new memory functions.
\section{Memory Model}
\subsection{Weighted Impact Memory Model (WIMM)}
Our WIMM hires the  decision function as
\begin{equation}\label{eq:w1}
  f(\text{x}) = <\text{w}, \varphi (\text{x})>+b+\sum\limits_{i = 1}^{m} y_i \xi_i\delta (\text{x}_i,\text{x}),
\end{equation}
where $\delta (\text{x}_i,\text{x})$ is the memory influence function, and it can be the similarity function between $\text{x}_i$ and $\text{x}$,
e.g.,
\begin{equation}\label{eq:mem1}
  \delta(\text{x}_i,\text{x}_j) = \frac{1}{\sigma \sqrt{2\pi}} \exp {(-\frac{\parallel  \text{x}_i -\text{x}_j \parallel  ^2}{2\sigma  ^2})},\quad \sigma >0,
\end{equation}
  
\begin{equation}\label{eq:mem2}
  \delta(\text{x}_i,\text{x}_j) = \max \{\rho - \parallel  \text{x}_i -\text{x}_j\parallel ,0 \},\quad \rho>0,
\end{equation}

\begin{equation}\label{eq:mem3}
  \delta(\text{x}_i,\text{x}_j) = 
  \begin{cases}
    \parallel  \text{x}_i -\text{x}_j\parallel,  \quad \text{if} \; \parallel  \text{x}_i -\text{x}_j\parallel \leq \varepsilon,\;\varepsilon>0, \\
      \; 0, \qquad \quad \; {\rm else}, \\
  \end{cases} \\
\end{equation}
and
\begin{equation}\label{eq:mem4}
  \delta(\text{x}_i,\text{x}_j) = \begin{cases}
      \frac{b}{\parallel  \text{x}_i -\text{x}_j\parallel },\quad {\rm if} \; \text{x}_i\neq, \text{x}_j\;b>0,\\
     \; 1, \qquad \quad {\rm else}.\\
  \end{cases} \\
\end{equation}

The above functions measure the similarity between $\text{x}_i$ and $\text{x}_j$. These influence functions are symmetric, and the memory of each training sample will have an effect on the prediction only if its memory cost is not zero. Then, when combined with the decision function (\ref{eq:w1}), the effect of memory can be achieved.

Therefore, our WIMM considers to
\begin{equation}\label{eq1}
  \begin{split}
    \min_{w,b,\xi } \quad 
      &\frac{1}{2}  \|\text{w} \|^2 + \frac{\gamma}{2}\sum_{i = 1}^{m} \xi _i^2 +\lambda \sum_{i = 1}^{m}\sum_{j = 1}^{m} y_iy_j \xi _j\delta (\text{x}_i,\text{x}_j),
      \\
      {\rm s.t.} \quad 
      &y_i(<\text{w},\varphi (\text{x}_i)>+b+\sum_{j = 1}^{m} y_j \xi _j\delta (\text{x}_i,\text{x}_j) )=1,  \qquad   \\ &i =1,...,m,
\end{split}
\end{equation}
where $\lambda ,\gamma$ is a positive parameter, $\xi=(\xi_1,...,\xi_m)^\top $ denotes the memory costs of training samples, and $\delta (x_i, x_j)$ is the memory impact function. Obviously, we use the decision function (\ref{eq:w1}), set the memory cost as a variable and predefine the memory influence function in the decision. From the constraints of (\ref{eq1}), it is necessary to remember all the training samples. The goal of problem (\ref{eq1}) is to find the optimal strategy with the lowest possible memory cost as well as memory impact.
To solve problem (\ref{eq1}), we derive its Lagrangian function as 
\begin{equation}
  \begin{split}
    L(\text{w},b,\xi)=\frac{1}{2}  \|\text{w} \|^2 + \frac{\gamma}{2}\sum_{i = 1}^{m} \xi _i^2
    +\lambda \sum_{i = 1}^{m}y_i\sum_{j = 1}^{m} y_j \xi _j\delta (\text{x}_i,\text{x}_j)
    \\ 
 +\sum_{i=1}^{m}\alpha _i(1-y_i(<\text{w},\varphi (\text{x}_i)>+b+\sum_{j = 1}^{m} y_j \xi _j\delta (\text{x}_i,\text{x}_j))),
  \end{split}
\end{equation}
where $\alpha_i \in \mathbb{R}$ is the Lagrangian multiplier with $i=1,\ldots,m$.
Let its partial derivatives w.r.t.  $ \text{w},b,\xi_i $ and $\alpha _i $ equal zeros, and we have
\begin{equation}
    \begin{cases}
        \frac{\partial L}{\partial \text{w}}=\text{w}-\sum\limits_{i=1}^{m}\alpha _iy_i\varphi (\text{x}_i),\\
        \frac{\partial L}{\partial b}=\sum\limits_{i = 1}^{m}\alpha _iy_i,\\
        \frac{\partial L}{\partial \xi_i }=c\xi _i+\lambda y_i\sum\limits _{j=1}^{m}y_j\delta (\text{x}_i,\text{x}_j)- y_i\alpha _i\sum\limits _{j=1}^{m}y_j\delta (\text{x}_i,x_j)=0,\\
        \frac{\partial L}{\partial \alpha _i }=1-y_i(<w,\varphi (\text{x}_i)>_i+b+\sum\limits_{j = 1}^{m} y_j \xi _j\delta (\text{x}_i,\text{x}_j)). \\
    \end{cases}
\end{equation}
Letting the partial derivative equal 0 gives
\begin{equation}\label{eq:w2}
  \begin{cases}
    \text{w}=\sum\limits_{i=1}^{m}\alpha _iy_i\varphi (\text{x}_i),\\
     \sum\limits_{i = 1}^{m}\alpha _iy_i=0,\\
     \xi _i=\frac{\alpha _iy_i\sum\limits _{j=1}^{m}y_j\delta (\text{x}_i,\text{x}_j)-\lambda y_i\sum\limits_{j=1}^{m}j\delta (\text{x}_i,\text{x}_j)}{c}, \quad i =1,\ldots,m,\\
    y_i(<w,\varphi (\text{x}_i)>_i+b+\sum\limits_{j = 1}^{m} y_j \xi _j\delta (\text{x}_i,\text{x}_j) )=1, \\
    ~~~~~~~~~~~~~~~~~~~~~~~~~~~~~~~~~~~~~~~~~~~~~~~~~~~~~~~~~~
       i=1,\ldots,m.
  \end{cases}
\end{equation}
To facilitate the solution, we reformulate problem (\ref{eq:w2}) as
\begin{equation}\label{eq:w3}
    \begin{pmatrix}
		\textbf{Y}K(\textbf{X},\textbf{X})\textbf{Y}+ \textbf{Y}\bigtriangleup \bigtriangleup^\top\textbf{Y} & \textbf{Y}\textbf{1} \\
		\textbf{1}^\top\textbf{Y} & 0
	\end{pmatrix}
    \begin{pmatrix}
		\alpha   \\
		b
	\end{pmatrix}=
    \begin{pmatrix}
      \textbf{1}+\frac{\lambda }{\gamma}\bigtriangleup \bigtriangleup^\top\textbf{1}\\
		0
	\end{pmatrix},
\end{equation}
where $\bigtriangleup \in \mathbb{R} ^{m \times m}$and its elements are $\delta(\text{x}_i,\text{x}_j) $ with $i,j=1,... ,m.$,, $K(\textbf{X},\textbf{X})$ is a kernel matrix, $\alpha =(\alpha _1,... ,\alpha _m)^\top$ and $\textbf{1}=(1,... ,1)^\top$. After solving the above system of equations, the final decision is
\begin{equation}
  f(\text{x}) = \sum\limits_{i = 1}^{m} y_i\alpha _i K(\text{x}_i,\text{x})+b+\sum\limits_{i = 1}^{m} y_i \xi_i\delta (\text{x}_i,\text{x}).
\end{equation}
Furthermore, by in problem (\ref{eq:w2}) we obtain $\xi =(\xi_1,... ,\xi_m)^\top$.
\subsection{Maximum Impact Memory Model (MIMM)}
Different from the WIMM, our MIMM selects the closest training sample of the unknown sample to affect it by decision function as
\begin{equation}\label{eq:m1}
    f(\text{x})=<\text{w}, \varphi (\text{x})>+b+y_i\xi _i\delta(\text{x},\text{x}_k),
\end{equation}
where $\text{x}_k$ is denoted as the centroid of the training point $\text{x}_{i}$ of the same kind.

For example, suppose
$\overline{x}_+$  and $\overline {x}_-$  are positive and the negative class centroids,
respectively. It is a straightforward way to use $x_+ $ or $x_-$ in $\delta(x_k,x)$ as the memory influence function. 

Thus, our MIMM considers to

\begin{equation}\label{eq2}
  \begin{split}
    \min_{\text{w},b,\xi } \quad 
      &\frac{1}{2}  \|\text{w} \|^2 + \frac{\gamma}{2}\sum_{i = 1}^{m} \xi _i^2 +\lambda \sum_{i = 1}^{m} \xi _i\delta _i  \\
      {\rm s.t.} \quad 
      &y_i(<\text{w},\varphi (\text{x}_i)>+b)=1-\xi _i\delta _i,  \qquad   i=1,\ldots,m,\\
\end{split}
\end{equation}
where $\delta _i=\delta (x_k,x_i)$ is the memory impact function we define.
Instead of using all training samples in our MIMM decision, as in WIMM, we memorize the training samples by finding the closest training samples to the test sample points. Correspondingly, the Lagrangian function of (\ref{eq2}) is
\begin{equation}
  \begin{split}
L(\text{w},b,\xi)=\frac{1}{2}  \|\text{w} \|^2 + \frac{\gamma}{2}\sum_{i = 1}^{m} \xi _i^2
 +\lambda \sum_{i = 1}^{m}\xi _i\delta _i
 \\
  +\sum_{i=1}^{m}\alpha _i(1-\xi _i\delta _i-y_i(<w,\varphi (\text{x}_i)>+b)).
 \end{split}
\end{equation}
Find the partial derivatives of w.r.t. $ w,b,\xi_i $and $\alpha _i $.,and we have
\begin{equation}
    \begin{cases}
        \frac{\partial L}{\partial \text{w}}=\text{w}-\sum\limits_{i=1}^{m}\alpha _iy_i\text{x}_i,\\
        \frac{\partial L}{\partial b}=\sum\limits_{i = 1}^{m}\alpha _iy_i,\\
        \frac{\partial L}{\partial \xi_i }=c\xi _i+\lambda \delta _i- \alpha _i\delta _i,\\
        \frac{\partial L}{\partial \alpha _i}= 1-\xi _i\delta _i-y_i(<w,\varphi (\text{x}_i)>+b)   \quad i=1,\ldots,m.
    \end{cases}
\end{equation}
Letting the partial derivative equal 0 gives
\begin{equation}\label{eq:m2}
  \begin{cases}
       \text{w}=\sum\limits_{i=1}^{m}\alpha _iy_i\text{x}_i,\\
       \sum\limits_{i = 1}^{m}\alpha _iy_i=0,\\
      \xi _i=\frac{\alpha _i\delta _i-\lambda \delta _i}{\gamma }, \quad i=1,\ldots,m,\\
       y_i(<w,\varphi (\text{x}_i)>+b)-1+\xi_i\delta _i=0,   \quad i=1,\ldots,m.
  \end{cases}
\end{equation}
After simplifying the system of equations, we get:
\begin{equation}\label{eq:m3}
  \begin{pmatrix}
  \textbf{Y}K(\textbf{X},\textbf{X})\textbf{Y}+ \textbf{Y}\textbf{D}\textbf{D}^\top\textbf{Y} & \textbf{Y}\textbf{1} \\
  \textbf{1}^\top\textbf{Y} & 0
\end{pmatrix}
  \begin{pmatrix}
  \alpha   \\
  b
\end{pmatrix}=
  \begin{pmatrix}
    \textbf{1}+\frac{\lambda }{\gamma}\textbf{D}\textbf{D}^\top \textbf{1}\\
  0
\end{pmatrix},
\end{equation}
where $\textbf{D}  \in \mathbb{R} ^{m \times m}$and its a diagonal matrix with $\textbf{D} _{ii}=\delta _i     (i=1,... ,m)$.

Thus, the WIMM model obtains $b$ and $\alpha$ by solving the system of linear equations (\ref{eq:m3}), and then $\xi$ by the optimality condition (\ref{eq:m2}), the final decision is

\begin{equation}
  f(\text{x}) = \sum\limits_{i = 1}^{m} y_i\alpha _i K(\text{x}_i,\text{x})+b+y_i\xi _i\delta _(\text{x},\text{x}_k).
\end{equation}

\begin{figure}[htbp]
	\centering
  \begin{minipage}{0.49\linewidth}
		\centering
    \captionsetup{justification=centering}	
		\includegraphics[width=4.5cm]{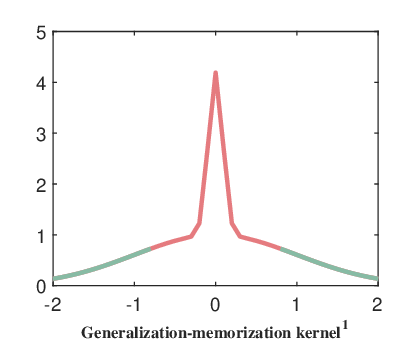}
	\end{minipage}
	\begin{minipage}{0.49\linewidth}
		\centering
		\includegraphics[width=4.5cm]{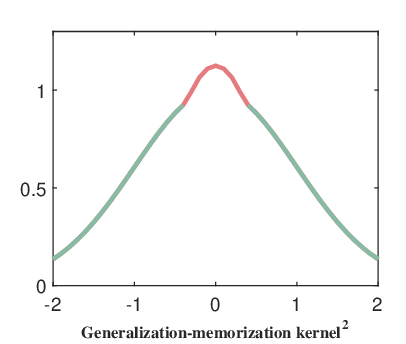}	
	\end{minipage}

	\begin{minipage}{0.49\linewidth}
		\centering
		\includegraphics[width=4.5cm]{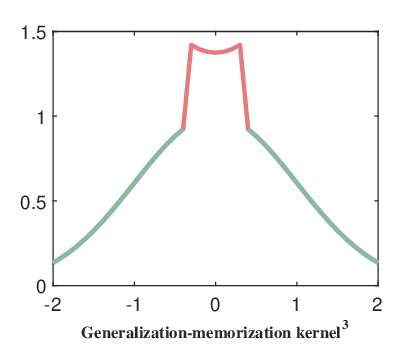}
	\end{minipage}
	\begin{minipage}{0.49\linewidth}
		\centering
		\includegraphics[width=4.5cm]{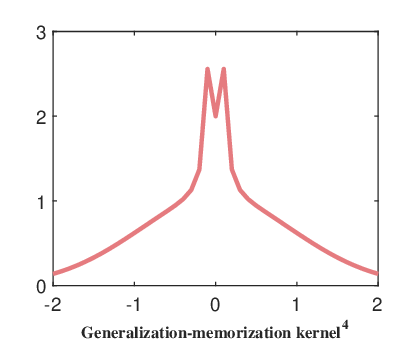}	
	\end{minipage}
  \caption{Different types of memory kernels.$^1,^2,^3$ and $^4$ with the influence function \ref{eq:mem1}, \ref{eq:mem2}, \ref{eq:mem3} and \ref{eq:mem4}. Where red indicates the extent of memory influence and green indicates the extent of generalization influence.}
	\label{Fig: mem}	
\end{figure}

Indeed, the advantage of memorization becomes evident when we combine the memorization function with the LSSVM method. This combination allows us to carefully observe and analyze the impact of each memorization influence function on the overall performance of the model of the combined model. Specifically, consider a learner that incorporates a generalized kernel $K_g(x, x) = \exp{(-\frac{\left\lVert x_i -x_j\right\rVert ^2}{\sigma ^2})}$ and a memory kernel $K_m$, where the memory influence function is chosen as equations (\ref{eq:mem1}), (\ref{eq:mem2}), (\ref{eq:mem3}), and (\ref{eq:mem4}).
Figure \ref{Fig: mem} illustrates the generated memory influence. With this memory influence, we can intuitively observe the range and degree of influence for each different influence function. We utilize the memory influence function to establish a rule, where classification is remembered only within a small region around the training data points. By adjusting the parameters of the influence function, we control the trade-off between generalization and memory in the algorithm.
\section{Discussion}
\newtheorem{Proposition}{Proposition}
\begin{Proposition}\label{th:1}
  The empirical risk of WIMM is zero if and only if problem (\ref{eq1}) has at least one feasible solution. Similarly, the empirical risk of MIMM is zero if and only if problem (\ref{eq2}) has at least one feasible solution.
\end{Proposition}
  The feasibility of problems (\ref{eq1}) or (\ref{eq2}) depends on the properties of the memory influence matrices $\bigtriangleup$ or \textbf{D}. Generally, we have the following sufficient conditions for practical applications. 
  \begin{Proposition}\label{th:2}
    The empirical risk of WIMM is zero if and only if matrix $\bigtriangleup $ is nonsingular. Similarly, the empirical risk of MIMM is zero if and only if the $\textbf{D} $ matrix is non-singular.
  \end{Proposition}
  Proof.
  We have considered the case where the $\bigtriangleup$ matrix is non-singular. It can be shown that $\textbf{Y}(K(\textbf{X},\textbf{X})+\bigtriangleup)\textbf{Y}^\top$ is also non-singular. Additionally, as $r(\textbf{1}^\top\textbf{Y})=1$, the problem (\ref{eq:w3}) must have a unique solution. Similarly, it can be demonstrated that when the $\textbf{D}$ matrix is non-singular, the problem (\ref{eq:m3}) must also have a unique solution. This conclusion follows from proposition (\ref{th:1}).
  $\hfill\square$

  \begin{Proposition}
    MIMM is equivalent to the LSSVM model if and only if $\textbf{D}$ is a unit array and $\lambda = 0$.
  \end{Proposition}
  
  Proof. 
  When $\textbf{D}$ is a unitary matrix and $\lambda = 0$, the problem (\ref{eq:m3}) is clearly in the form of a least squares system of linear equations to be solved. The proposition is proved and the conclusion holds.
  $\hfill\square$
  \begin{Proposition}
    WIMM is equivalent to the LSSVM model if and only if $\bigtriangleup$ is a unit array and $\lambda = 0$.
  \end{Proposition}
  Proof.
  When $\bigtriangleup$ is a unitary matrix and $\lambda = 0$, the problem (\ref{eq:w3}) is clearly in the form of a system of linear equations solved by least squares. The proposition is proved and the conclusion holds.
  $\hfill\square$

\begin{figure}
    \centering
      \includegraphics[scale=0.36]{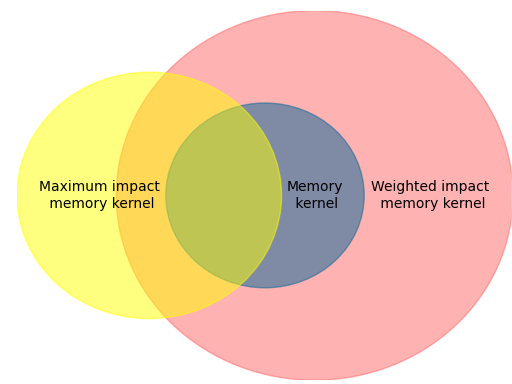}
    \caption{A memory relation diagram for MIMM,WIMM and SVM$^m$\cite{2021Reinforced}, where yellow, pink and blue-gray denote the $\textbf{D}$, $\bigtriangleup$ and $k_m$ memory kernel matrices, respectively.}
    \label{fig:image} 
  \end{figure}

  Fig. (\ref{fig:image}) gives information about the interrelationships between the three memory kernels by comparing the memory kernels of equations (\ref{eq:m3}), (\ref{eq:w3}) and SVM$^m$\cite{2021Reinforced}. We can find that these three memory kernels are $\textbf{D}$, $\bigtriangleup$ and $k_m$, which can be obtained by the matrix structure, $\textbf{D}$ is a diagonal matrix, and $\bigtriangleup$ and $k_m$ are symmetric matrices. By tuning the parameters, $\textbf{D}$, $\bigtriangleup$ and $k_m$ can be varied to a unitary matrix. Thus there exists an intersection of these three memory kernels. Since $\bigtriangleup$ can choose more than just one type of Gaussian kernel, $\bigtriangleup$ contains $k_m$. Since $\bigtriangleup$ and $\textbf{D}$ have different influence functions, $\bigtriangleup$ and $\textbf{D}$ only have an intersection but no containment relationship.
\section{Experiments}
This section utilizes several calibration datasets from UCI, for which Table (\ref{tab:1}) provides detailed information. We analyze the performance of our WIMM and MIMM models on various benchmark datasets, along with their execution times on large datasets. Additionally, we test the generalization performance of the two models and their ability to adapt to noise.
The classical LSSVM utilizes linear kernels, while the SVM$^m$ and HGMM models employ linear generalization kernels and RBF memory kernels. In contrast, our WIMM and MIMM models both utilize linear kernels. All these models were implemented using MATLAB 2017a on a PC equipped with an Intel Core Duo processor (dual 4.2 GHz) and 32 GB of RAM.
For the RBF kernel $K(x_i,x_j)= \exp(-\sigma \parallel x_i-x_j\parallel ^2)$, we tested parameters $\sigma$ from the set ${2^i | i = -6, -5, ..., 5}$, and for other models, we tested weighing parameters from the same set.
To begin the comparison, we evaluated the memory performance of the linear kernel in WIMM and MIMM models on some small datasets, with the linear kernel in LSSVM used as the benchmark.
\begin{table}[!t]
  \caption{Details of benchmark datasets\label{tab:1}}
  \centering
    \begin{tabular}{cccc}
      \toprule
     ID & Name & m & n \\
      \midrule
      (a) & Cleveland & 173 & 13 \\
      (b) & Ionosphere & 351 & 34 \\
      (c) & New-thyroid & 215 & 4 \\
      (d) & Parkinsons & 195 & 22 \\
      (e) & Sonar & 208 & 60 \\
      (f) & TicTacToe & 958 & 27 \\
      (g) & Vowel & 988 & 13 \\
      (h) & Wisconsin & 683 & 9 \\
      (i) & German & 1000 & 20 \\
      (j) & Shuttle & 1829 & 9 \\
      (k) & Segment & 2308 & 19 \\
      (l) & Waveform & 5000 & 21 \\
      (m) & TwoNorm & 7400 & 20 \\
      (n) & IJCNN01 & 49990 & 22 \\
      \bottomrule
  \end{tabular}
  \end{table}

  To assess the memory capacity of the WIMM model, Table (\ref{tab:2}) presents the highest training and testing accuracies achieved by the WIMM model. This table provides valuable insights into the model's ability to memorize and generalize effectively on the tested datasets. It can be observed from Table (\ref{tab:2}) that the WIMM model with the memory influence function (\ref{eq:mem1}, \ref{eq:mem2}, \ref{eq:mem3}) achieves a training accuracy of $100\%$ on all datasets. However, the failure to reach $100\%$ training accuracy can be attributed to the irreversibility of the $\bigtriangleup$ term in the function and the impact of different influence functions on the data's memory capacity.
Among the various influence functions, the memory influence function (\ref{eq:mem1}) yields the highest test accuracy for most of the datasets. Consequently, for the remaining experiments, we utilize the memory influence function (\ref{eq:mem1}) as the basis for our WIMM model.
\begin{table*}[ht]
  \caption{Testing and training accuracy of WIMM and LSSVM using memory effects.}
  \label{tab:2}
  \begin{tabular}{p{0.2cm}p{1.49cm}p{1cm}p{1cm}p{1cm}p{1.49cm}|p{1.49cm}p{1.49cm}p{1.49cm}p{1.49cm}p{1.49cm}p{1.49cm}}
    \toprule
    ID &LSSVM & WIMM$^1$&WIMM$^2$ & WIMM$^3$ & WIMM$^4$ & LSSVM & WIMM$^1$ & WIMM$^2$ & WIMM$^3$ & WIMM$^4$ \\
    & train($\%$) & train($\%$) & train($\%$)& train($\%$)&train($\%$)& test($\%$)& test($\%$)& test($\%$)& test($\%$)& test($\%$)\\
    \toprule
    (a) & $96.39\pm0.47$ & $100\pm0$ & $100\pm0$ & $100\pm0$ & $90.77\pm1.0$ & $94.82\pm4.18$ & $95.44\pm3.36$ & $95.36\pm3.29$ &\pmb{$95.89\pm4.49$} & $92.85\pm6.16$ \\
    (b) & $89.46\pm0.47$ & $100\pm0$ & $100\pm0$ & $100\pm0$ & $44.02\pm1.6$ & $88.3\pm3.46$ & $88.31\pm3.13$ & $88.36\pm5.52$ & \pmb{$89.77\pm3.86$} & $42.24\pm9.23$ \\
    (c) & $94.08\pm0.86$ & $100\pm0$ & $100\pm0$ & $100\pm0$ & $72.9\pm2.08$ & $93.66\pm3.41$ &\pmb{$97.79\pm2.15$}& $87.94\pm4.84$ & $89.11\pm7.7$ & $88.31\pm4.52$ \\
    (d) & $91.42\pm1.15$ & $100\pm0$ & $100\pm0$ & $100\pm0$ & $64.57\pm5.01$ & $88.4\pm7.44$ & \pmb{$94.73\pm4.95$ }& $88.7\pm5.84$ & $91.8\pm3.17$ & $83.9\pm4.3$ \\
    (e) & $87.99\pm1.36$ & $100\pm0$ & $100\pm0$ & $100\pm0$ & $83.65\pm1.4$ & $79.48\pm2.85$ &\pmb{$86.79\pm8.43$} & $78.89\pm1.12$ & $80.29\pm5.84$ & $79.03\pm8.03$ \\
   (f) & $98.33\pm0.25$ & $100\pm0$ & $100\pm0$ & $100\pm0$ & $100\pm0$ & \pmb{$98.33\pm1.0$ }& \pmb{$98.33\pm1.13$ }& \pmb{$98.33\pm1.62$ }& \pmb{$98.33\pm1.13$ }& $65.35\pm4.38$ \\
    (g) & $95.04\pm0.34$ & $100\pm0$ & $100\pm0$ & $100\pm0$ & $75.05\pm7.08$ & $95.04\pm2.08$ & \pmb{$100\pm0$} & $99.8\pm0.28$ & $99.8\pm0.45$ & $94.84\pm1.08$ \\
    (h) & $96.16\pm0.61$ & $100\pm0$ & $100\pm0$ & $100\pm0$ & $95.65\pm0.71$ & $96.18\pm2.45$ & $96.63\pm1.52$ & $96.65\pm2.25$ & \pmb{$96.78\pm1.84$}& $90.05\pm3.13$ \\
    \bottomrule
    \multicolumn{8}{l}{\small $^1,^2,^3$ and $^4$ with the influence function \ref{eq:mem1}, \ref{eq:mem2}, \ref{eq:mem3} and \ref{eq:mem4}.}\\
  \end{tabular}
\end{table*}

\begin{table*}[htbp]
  \footnotesize
  \centering
  \caption{Testing and training accuracy of MIMM and LSSVM using memory effects.}
  \label{tab:3}
  \begin{tabular}{p{0.2cm}p{1.5cm}p{1.4cm}p{1.5cm}p{1.4cm}p{0.9cm}|p{1.5cm}p{1.5cm}p{1.5cm}p{1.5cm}p{1.5cm}p{1.5cm}}
    \toprule
ID &LSSVM & MIMM$^1$ & MIMM$^2$ & MIMM$^3$ & MIMM$^4$ & LSSVM & MIMM$^1$ & MIMM$^2$ & MIMM$^3$ & MIMM$^4$\\ %
    & train($\%$) & train($\%$) & train($\%$)& train($\%$)&train($\%$)& test($\%$)& test($\%$)& test($\%$)& test($\%$)& test($\%$)\\ 
    \midrule
(a) & $96.39\pm0.47$ & $96.4\pm0.79$ & $94.83\pm3.1$ & $98.25\pm0.9$ & $100\pm0$ & $94.82\pm4.18$ & $95.28\pm3.4$ & $94.72\pm4.81$ & $95.33\pm1.71$ & \pmb{$95.36\pm2.6$} \\
(b) & $89.46\pm0.47$ & $100\pm0$ & $91.45\pm2.01$ & $100\pm0$ & $100\pm0$ & $88.3\pm3.46$ & $90.08\pm4.29$ & \pmb{$94.61\pm3.64$} & $88.61\pm1.67$ & $89.75\pm4.63$ \\
(c) & $94.08\pm0.86$ & $100\pm0$ & $99.09\pm2.03$ & $100\pm0$ & $100\pm0$ & $93.66\pm3.41$ & $98.64\pm1.18$ & $98.57\pm1.28$ & $98.72\pm1.9$ &\pmb{ $98.73\pm1.9$} \\
(d) & $91.42\pm1.15$ & $100\pm0$ & $97.64\pm2.25$ & $100\pm0$ & $100\pm0$ & $88.4\pm7.44$ & \pmb{$97.49\pm1.49$} & $97.07\pm1.93$ & $96.42\pm3.03$ & $96.3\pm1.62$ \\
(e) & $87.99\pm1.36$ & $100\pm0$ & $86.85\pm2.73$ & $100\pm0$ & $100\pm0$ & $79.48\pm2.85$ & $86.11\pm5.32$ & $87.46\pm3.01$ & $86.94\pm2.2$ & \pmb{$88.47\pm2.63$} \\
(f) & $98.33\pm0.25$ & $100\pm0$ & $98.33\pm1.44$ & $100\pm0$ & $100\pm0$ & \pmb{$98.33\pm1.0$}& \pmb{$98.33\pm1.7$} &\pmb{ $98.33\pm0.93$} & \pmb{$98.33\pm0.93$} & \pmb{$98.33\pm1.13$ }\\
(g) & $95.04\pm0.34$ & $100\pm0$ & $100\pm0$ & $100\pm0$ & $100\pm0$ & $95.04\pm2.08$ & \pmb{$100\pm0$ }& \pmb{$100\pm0$} & \pmb{$100\pm0$} & \pmb{$100\pm0$} \\
(h) & $96.16\pm0.61$ & $100\pm0$ & $97.23\pm1.29$ & $100\pm0$ & $100\pm0$ & $96.18\pm2.45$ & $97.36\pm1.33$ &\pmb{$97.37\pm0.82$} & $97.08\pm1.0$ & $96.94\pm1.65$ \\
    \bottomrule
    \multicolumn{8}{l}{\small $^1,^2,^3$ and $^4$ with the influence function \ref{eq:mem1}, \ref{eq:mem2}, \ref{eq:mem3} and \ref{eq:mem4}.}\\
  \end{tabular}
\end{table*}

Likewise, to evaluate the memory capacity of the MIMM model, Table (\ref{tab:3}) displays the maximum training and testing accuracies achieved by the MIMM model.  It is evident from Table (\ref{tab:3}) that the MIMM model attains a training accuracy of $100\%$ when using the memory influence function (\ref{eq:mem4}).
The reason the other influence functions do not achieve $100\%$ training accuracy is due to the irreversibility of $\textbf{D}$ in these functions.
The choice of different influence functions impacts the data's memory capacity.
Among the various influence functions, the memory influence function (\ref{eq:mem4}) yields the highest test accuracy for the majority of the datasets. As a result, for the subsequent experiments, we adopt the memory influence function (\ref{eq:mem4}) as the basis for our MIMM model.

Next, to compare the running times of other memory models under optimal parameters, we recorded the execution times along with the corresponding accuracies on a larger dataset. This evaluation allows us to further assess the trade-off between the time consumed for memorization and the achieved performance on the same task for different memory models. 
For each dataset, approximately $70\%$ of the total samples were randomly selected for training, ensuring that half of them belonged to the positive category and the other half to the negative category, while the remaining samples constituted the test set. This process was repeated five times, and the highest average training accuracy along with its standard deviation, the corresponding highest test accuracy, and the time taken to run the model once with optimal parameters were recorded for each dataset. The shortest time spent is indicated in bold in Table (\ref{tab:4}).
From Table (\ref{tab:4}), it is evident that the test accuracies do not differ significantly. Notably, both WIMM and MIMM exhibit shorter execution times compared to HGMM and SVM$^m$. This efficiency can be attributed to the fact that WIMM and MIMM models are solved as linear system of equations, whereas HGMM and SVM$^m$ are solved as quadratic programming problems.

\begin{figure*}[htbp!]  
  \centering
  \begin{minipage}{0.35\linewidth}
    \centering
    \captionsetup{justification=centering}	
    \includegraphics[width=6cm]{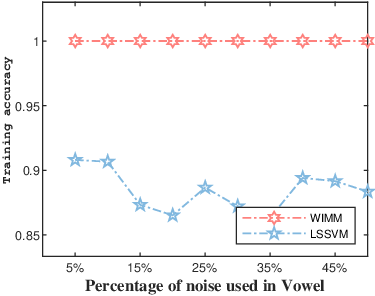}
  \end{minipage}
  \begin{minipage}{0.45\linewidth}
    \centering
    \captionsetup{justification=centering}	
    \includegraphics[width=6cm]{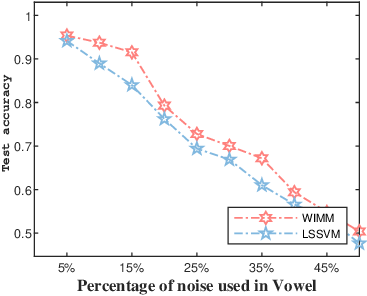}
  \end{minipage}
  \begin{minipage}{0.35\linewidth}
    \centering
    \captionsetup{justification=centering}
    \includegraphics[width=6cm]{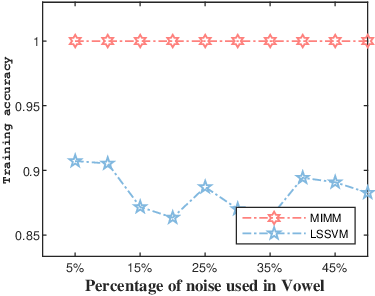}
  \end{minipage}
  \begin{minipage}{0.45\linewidth}
    \centering
    \captionsetup{justification=centering}	
    \includegraphics[width=6cm]{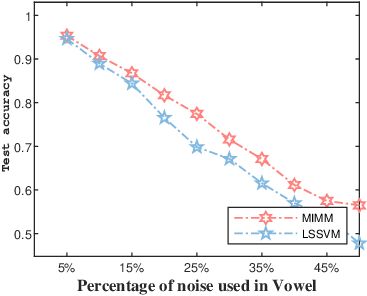}
  \end{minipage}
    \caption{Training (left)/Testing (right) accuracy at different noise points}
  \label{Fig:noise1}	
\end{figure*}

\begin{figure*}[htbp!]  
  \centering
  \begin{minipage}{0.35\linewidth}
    \centering
    \captionsetup{justification=centering}
    \includegraphics[width=6cm]{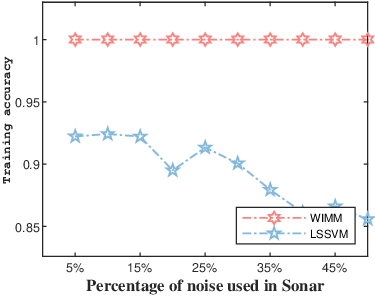}
  \end{minipage}
  \begin{minipage}{0.45\linewidth}
    \centering
    \captionsetup{justification=centering}
    \includegraphics[width=6cm]{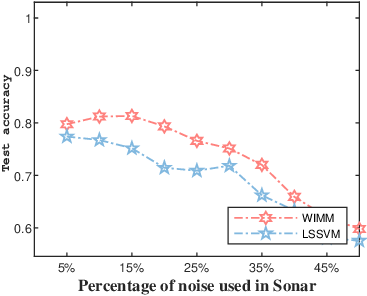}
  \end{minipage}
  \begin{minipage}{0.35\linewidth}
    \centering
    \captionsetup{justification=centering}
    \includegraphics[width=6cm]{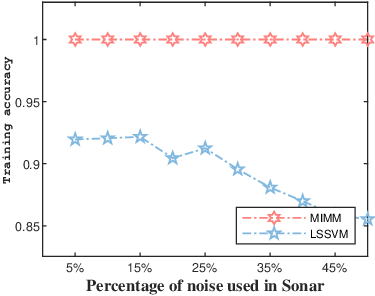}
  \end{minipage}
  \begin{minipage}{0.45\linewidth}
    \centering
    \captionsetup{justification=centering}
    \includegraphics[width=6cm]{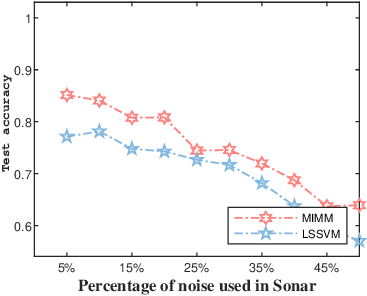}
  \end{minipage}
  \caption{Training (left)/Testing (right) accuracy at different noise points}
  \label{Fig:noise2}	
\end{figure*}

  \begin{table*}[htbp]
    \centering
    \caption{Accuracy and time to train and test linear classifiers on benchmark datasets.}
    \label{tab:4}
    \begin{tabular}{p{0.5cm}p{1cm}p{1cm}p{1cm}p{1cm}|p{1.49cm}p{1.49cm}p{1.49cm}p{1.49cm}}
      \toprule
ID & SVM$^m$ & HGMM & WIMM & MIMM & SVM$^m$ & HGMM & WIMM & MIMM\\  
& train($\%$) & train($\%$)& train($\%$)&train($\%$)& test($\%$)& test($\%$)& test($\%$)& test($\%$)\\   & & &  &  & time(s)& time(s) & time(s) & time(s) \\
      \midrule
      (i) & $100\pm0$ & $100\pm0$ & $100\pm0$ & $100\pm0$ & $76.1\pm1.77$ & $78.33\pm3.53$ & $76.64\pm1.43$ & $72.06\pm2.55$ \\
      & & & & & 0.198s & 0.196s & 0.119s & \bf{0.105s}  \\
      (j) & $100\pm0$ & $100\pm0$ & $100\pm0$ & $100\pm0$ & $99.95\pm0.12$ & $100\pm0$ & $99.96\pm0.08$ & $100\pm0$ \\
      & & & & & 0.738s & 0.663s & 0.344s & \bf{0.291s} \\
      (h) & $100\pm0$ & $100\pm0$ & $100\pm0$ & $100\pm0$ & $99.83\pm0.18$ & $99.88\pm0.19$ & $99.86\pm0.14$ & $99.77\pm0.21$ \\
      & & & & & 0.997s & 1.256s & 0.652s & \bf{0.393s} \\
      (l) & $100\pm0$ & $100\pm0$ & $100\pm0$ & $100\pm0$ & $90.16\pm0.81$ & $88.27\pm0.61$ & $86.42\pm0.85$ & $83.24\pm0.67$ \\
      & & & & & 5.835s & 6.682s & 2.362s & \bf{1.656s} \\
      (m) & $100\pm0$ & $100\pm0$ & $100\pm0$ & $100\pm0$ & $98.02\pm0.25$ & $97.98\pm0.26$ & $97.99\pm0.5$ & $95.09\pm0.56$ \\
      & & & & & 15.793s & 18.531s & 5.888s & \bf{3.895s} \\
      (n) &\centering{$*$} & \centering{$*$} &$100\pm0$ &$100\pm0$ & \centering{$*$} & \centering{$*$} & $93.80\pm0.26$& $97.37\pm0.08$\\
      & & & & & & & 787.645s & \bf{435.870s} \\
      \bottomrule
      \multicolumn{8}{l}{\small '$*$' indicates a lack of sufficient operating memory.}\\
    \end{tabular}
  \end{table*} 
  In practical applications, many tasks involve learning with labeled noise. Therefore, to examine the ability of WIMM and MIMM models to adapt to noise, we conducted experiments with datasets containing labeled noise. 
For certain datasets from Table (\ref{tab:1}), we randomly select $80\%$ of the training samples to form the training set, while the remaining samples constitute the test set. We then introduce label noise to the training set, setting $5\%$, $10\%$, $15\%$, and gradually up to $50\%$ of the labels to the opposite class. This process is repeated five times, and we record the highest test accuracy along with the corresponding average training accuracy for comparison with LSSVM.
From Figures (\ref{Fig:noise1}) and (\ref{Fig:noise2}), we can observe the following trends: 
i) The training accuracy of LSSVM is not consistently $100\%$ except for our WIMM and MIMM models.
ii) The test performance of LSSVM is consistently lower than our model and exhibits instability with increasing label noise.
iii) The test performance of our models (WIMM and MIMM) shows a gradual decline as the label noise increases in a regular manner.
These observations suggest that our WIMM and MIMM models outperform LSSVM in handling labeled noise and offer more stable and robust performance under noisy conditions.

\begin{figure*}[htbp!]  
    \centering
    \begin{minipage}{0.35\linewidth}
      \centering
      \captionsetup{justification=centering}	
      \includegraphics[width=6cm]{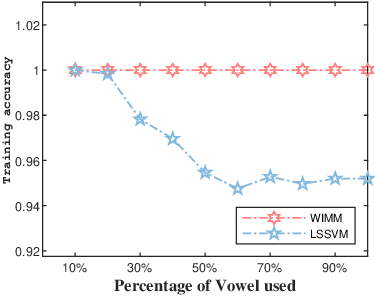}
    \end{minipage}
    \begin{minipage}{0.45\linewidth}
      \centering
      \includegraphics[width=6cm]{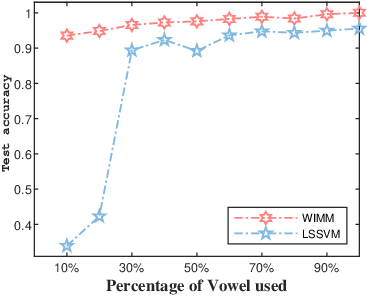}
    \end{minipage}
    \begin{minipage}{0.35\linewidth}
      \centering
      \includegraphics[width=6cm]{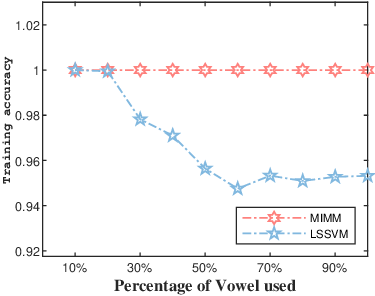}
    \end{minipage}
    \begin{minipage}{0.45\linewidth}
      \centering
      \includegraphics[width=6cm]{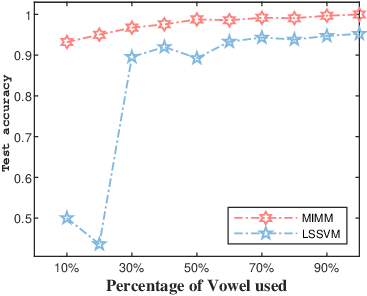}
    \end{minipage}
    \caption{Training(left)/Test(right) accuracy for different sample sizes.}
      \label{Fig:rang1}	
  \end{figure*}

\begin{figure*}[htbp!]  
    \centering
    \begin{minipage}{0.35\linewidth}
          \centering
          \captionsetup{justification=centering}	
      \includegraphics[width=6cm]{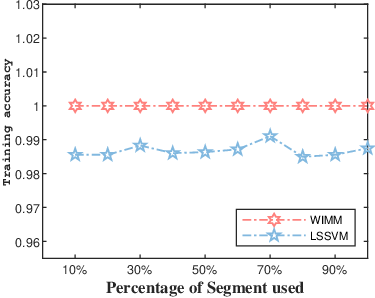}
    \end{minipage}
    \begin{minipage}{0.45\linewidth}
      \centering
      \captionsetup{justification=centering}	
      \includegraphics[width=6cm]{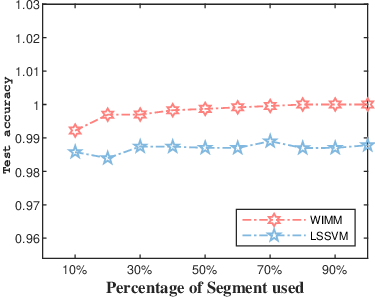}
    \end{minipage}
    \begin{minipage}{0.35\linewidth}
      \centering
      \captionsetup{justification=centering}	
      \includegraphics[width=6cm]{  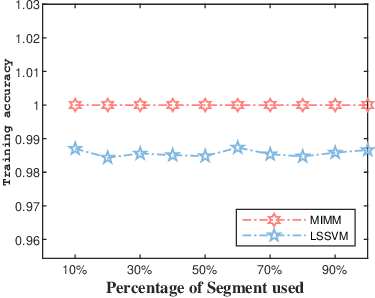}
    \end{minipage}
    \begin{minipage}{0.45\linewidth}
      \centering
      \captionsetup{justification=centering}	
      \includegraphics[width=6cm]{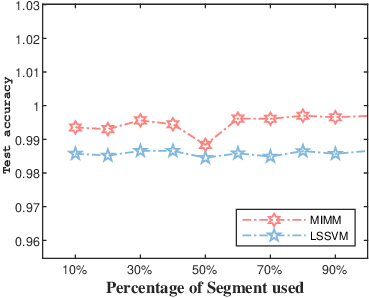}
  \end{minipage}
      \caption{Training(left)/Test(right) accuracy for different sample sizes.}
      \label{Fig:rang2}	
  \end{figure*}

Moreover, in many tasks, obtaining an adequate number of training samples can be particularly challenging. Hence, we further investigate the performance of these models in comparison to our WIMM and MIMM models under conditions of limited training samples. For selected datasets from Table (\ref{tab:1}), we randomly select $80\%$ of the samples to form the training set, and the remaining samples constitute the test set. Subsequently, we vary the proportion of training samples used, ranging from $10\%$ to $100\%$, in incremental steps. The models are tested on the dataset, and this process is repeated five times. We record the highest test accuracy along with the corresponding average training accuracy for comparison with LSSVM.
  From Figures (\ref{Fig:rang1}) and (\ref{Fig:rang2}), the following observations are made:
  i) Apart from our WIMM and MIMM models, the training accuracy of LSSVM is not consistently $100\%$.
ii) The test performance of LSSVM is consistently inferior to our models, and its performance improves as more training data is used.
  iii) The test performance of our models (WIMM and MIMM) demonstrates a steady improvement as the number of training samples increases.
  These findings suggest that our WIMM and MIMM models outperform LSSVM, especially when training data is limited, and they consistently achieve higher test accuracy as the number of training samples grows.
  
\begin{figure*}[htbp]
  \centering
  \begin{minipage}{0.35\linewidth}
    \centering
    \captionsetup{justification=centering}	
    \includegraphics[width=6cm]{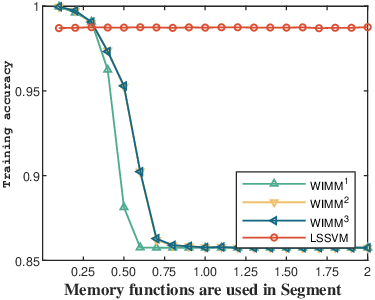}
  \end{minipage}
  \begin{minipage}{0.45\linewidth}
    \centering
    \includegraphics[width=6cm]{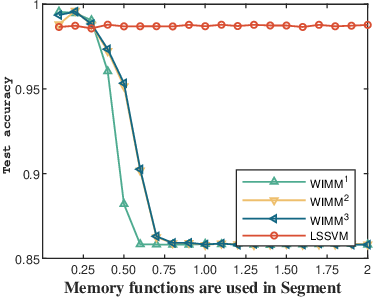}	
  \end{minipage}

  \begin{minipage}{0.35\linewidth}
    \centering
    \includegraphics[width=6cm]{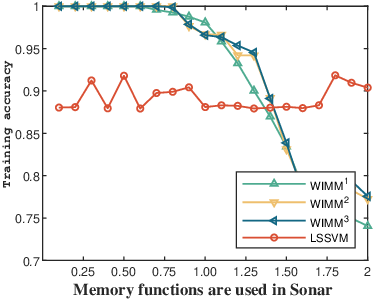}
  \end{minipage}
  \begin{minipage}{0.45\linewidth}
    \centering
    \includegraphics[width=6cm]{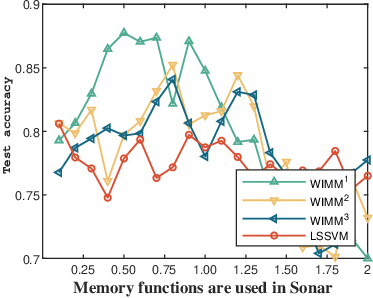}	
  \end{minipage}
  \caption{Training (left)/testing (right) accuracy with different influence functions.}
  \label{Fig:parameter1}	
\end{figure*}

\begin{figure*}[htbp]
  \centering
  \begin{minipage}{0.35\linewidth}
    \centering
    \captionsetup{justification=centering}	
    \includegraphics[width=6cm]{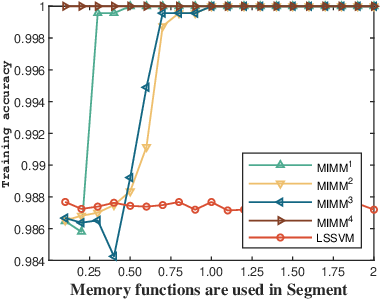}
  \end{minipage}
  \begin{minipage}{0.45\linewidth}
    \centering
    \includegraphics[width=6cm]{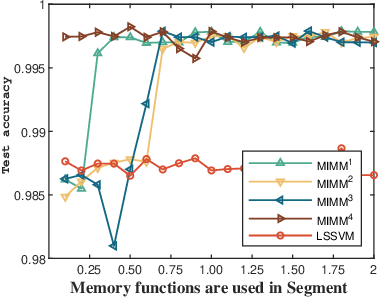}	
  \end{minipage}

  \begin{minipage}{0.35\linewidth}
    \centering
    \includegraphics[width=6cm]{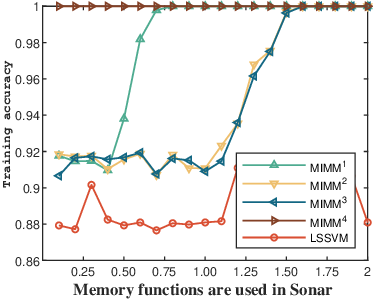}
  \end{minipage}
  \begin{minipage}{0.45\linewidth}
    \centering
    \includegraphics[width=6cm]{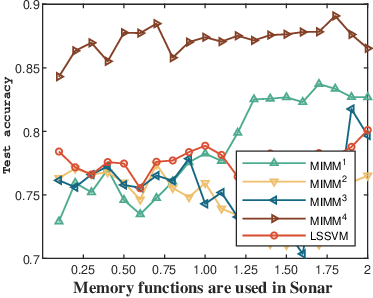}	
  \end{minipage}
  \caption{Training (left)/testing (right) accuracy with different influence functions.}	
  \label{Fig:parameter2}	
\end{figure*}

To compare the impact of different memory influence functions on the WIMM and MIMM models, we analyze the effect of memory kernel parameters on the models. Figures (\ref{Fig:parameter1}) and (\ref{Fig:parameter2}) illustrate this impact, with the LSSVM model used as a benchmark for comparison. In this experiment, we consider the models with parameters ranging from $\{0.1, 0.2, ..., 2\}$. Specifically, we focus on the Segment and Sonar datasets from Table (\ref{tab:1}). For these datasets, $80\%$ of the training samples are randomly selected as the training set, and the remaining samples form the test set. The process is repeated five times, recording the highest test accuracy and its corresponding average training accuracy for comparison with LSSVM. As the WIMM model with the memory influence function (\ref{eq:mem4}) demonstrated poorer results in Table (\ref{tab:2}), we consider the influence case of this model.
From Figures (\ref{Fig:parameter1}) and (\ref{Fig:parameter2}), we can make the following observations:
i) The WIMM model is more sensitive to the parameters, and its ability to memorize is contingent on selecting the appropriate parameters.
ii) The MIMM model, particularly with the memory influence function (\ref{eq:mem4}), exhibits greater stability, consistently memorizing the training samples while ensuring that the test performance remains superior to that of the LSSVM model.
iii) Overall, our models consistently outperform the LSSVM model, provided that we select the right parameters.
These findings emphasize the importance of parameter selection in the WIMM model, while the MIMM model offers a more robust performance with the chosen memory influence function. In general, our models demonstrate superior performance compared to the LSSVM model when the appropriate parameters are employed.

\section{Conclusion}
We have presented two novel innovations in the traditional LSSVM framework:
(i) We have proposed a replacement for the objective function of LSSVM, leading to improved performance.
(ii) We have introduced a new memory generalization kernel that effectively incorporates the complete memory of the training data, achieving zero training error.
As a result of these innovations, the MIMM and WIMM models have demonstrated superior generalization accuracy while maintaining the same computational complexity. Specifically, they still have involved solving a system of linear equations with a corresponding dimension, just like the current LSSVM implementation.
Furthermore, our models have exhibited higher classification accuracy and have enhanced noise tolerance on certain datasets. Additionally, they have required less time and have cost to memorize training samples compared to existing memory models.

In future work, we plan to extend our memory enhancement mechanism to other models and explore its applicability to a variety of other problems.
In addition, we intend to consider multiple memory patterns in our memory model and introduce forgetting mechanisms to enrich the memory capacity to effectively solve a wider range of tasks.
\section*{Acknowledgement}
This work is 
supported in part by National Natural Science Foundation of China (Nos. 12271131, 62106112, 61866010,61966024 and 11871183), in 
part by the Natural 
Science Foundation 
of Hainan Province 
(No.120RC449), and 
in part by the Key 
Laboratory of 
Engineering Modeling 
and Statistical 
Computation of 
Hainan Province.
\bibliographystyle{IEEEtran}
\bibliography{cas-refs}

\begin{thebibliography}{10}
\providecommand{\url}[1]{#1}
\csname url@samestyle\endcsname
\providecommand{\newblock}{\relax}
\providecommand{\bibinfo}[2]{#2}
\providecommand{\BIBentrySTDinterwordspacing}{\spaceskip=0pt\relax}
\providecommand{\BIBentryALTinterwordstretchfactor}{4}
\providecommand{\BIBentryALTinterwordspacing}{\spaceskip=\fontdimen2\font plus
\BIBentryALTinterwordstretchfactor\fontdimen3\font minus
  \fontdimen4\font\relax}
\providecommand{\BIBforeignlanguage}[2]{{%
\expandafter\ifx\csname l@#1\endcsname\relax
\typeout{** WARNING: IEEEtran.bst: No hyphenation pattern has been}%
\typeout{** loaded for the language `#1'. Using the pattern for}%
\typeout{** the default language instead.}%
\else
\language=\csname l@#1\endcsname
\fi
#2}}
\providecommand{\BIBdecl}{\relax}
\BIBdecl

\bibitem{memorization}
S.~Chatterjee, ``Learning and memorization,'' pp. 755--763, 2018.

\bibitem{10.1145/3357713.3384290}
V.~Feldman, ``Does learning require memorization? a short tale about a long
  tail,'' in \emph{Proceedings of the 52nd Annual ACM SIGACT Symposium on
  Theory of Computing}, 2020, pp. 954--959.

\bibitem{VAPNIK2021108018}
V.~Vapnik and R.~Izmailov, ``Reinforced svm method and memorization
  mechanisms,'' \emph{Pattern Recognition}, vol. 119, p. 108018, 2021.

\bibitem{1995Support}
C.~Cortes and V.~Vapnik, ``Support-vector networks,'' \emph{Machine learning},
  vol.~20, pp. 273--297, 1995.

\bibitem{2017A}
D.~Arpit \emph{et~al.}, ``A closer look at memorization in deep networks,'' in
  \emph{Proceedings of the 34th International Conference on Machine Learning},
  vol.~70, 2017, pp. 233--242.

\bibitem{cohen2019dnn}
G.~Cohen \emph{et~al.}, ``Dnn or k-nn: That is the generalize vs. memorize
  question,'' \emph{arXiv preprint arXiv:1805.06822}, 2018.

\bibitem{1990Finding}
J.~L. Elman, ``Finding structure in time,'' \emph{Cognitive science}, vol.~14,
  no.~2, pp. 179--211, 1990.

\bibitem{pmlr-v119-yang20j}
Z.~Yang \emph{et~al.}, ``Rethinking bias-variance trade-off for generalization
  of neural networks,'' in \emph{International Conference on Machine
  Learning}.\hskip 1em plus 0.5em minus 0.4em\relax PMLR, 2020, pp.
  10\,767--10\,777.

\bibitem{hinton2015distilling}
G.~Hinton, O.~Vinyals, and J.~Dean, ``Distilling the knowledge in a neural
  network,'' \emph{arXiv preprint arXiv:1503.02531}, 2015.

\bibitem{Goodfellow-et-al-2016}
I.~Goodfellow, Y.~Bengio, and A.~Courville, \emph{Deep learning}.\hskip 1em
  plus 0.5em minus 0.4em\relax MIT press, 2016.

\bibitem{bishop1995neural}
C.~M. Bishop, \emph{Neural networks for pattern recognition}.\hskip 1em plus
  0.5em minus 0.4em\relax Oxford university press, 1995.

\bibitem{1997Long}
S.~Hochreiter and J.~Schmidhuber, ``Long short-term memory,'' \emph{Neural
  computation}, vol.~9, no.~8, pp. 1735--1780, 1997.

\bibitem{10.1145/3446776}
C.~Zhang \emph{et~al.}, ``Understanding deep learning (still) requires
  rethinking generalization,'' \emph{Communications of the ACM}, vol.~64,
  no.~3, pp. 107--115, 2021.

\bibitem{Yang2023ResMemLW}
Z.~Yang \emph{et~al.}, ``Resmem: Learn what you can and memorize the rest,''
  \emph{arXiv preprint arXiv:2302.01576}, 2023.

\bibitem{2020WanShanshan}
S.~Wan and Z.~Niu, ``A hybrid e-learning recommendation approach based on
  learners’ influence propagation,'' \emph{IEEE Transactions on Knowledge and
  Data Engineering}, vol.~32, no.~5, pp. 827--840, 2019.

\bibitem{2020CullyAntoine}
A.~Cully and Y.~Demiris, ``Online knowledge level tracking with data-driven
  student models and collaborative filtering,'' \emph{IEEE Transactions on
  Knowledge and Data Engineering}, vol.~32, no.~10, pp. 2000--2013, 2019.

\bibitem{clayton2001elements}
N.~S. Clayton \emph{et~al.}, ``Elements of episodic--like memory in animals,''
  \emph{Philosophical Transactions of the Royal Society of London. Series B:
  Biological Sciences}, vol. 356, no. 1413, pp. 1483--1491, 2001.

\bibitem{PMID:26400190}
A.~N. Rafferty \emph{et~al.}, ``Faster teaching via pomdp planning,''
  \emph{Cognitive science}, vol.~40, no.~6, pp. 1290--1332, 2016.

\bibitem{8481496}
Q.~Kang and W.~P. Tay, ``Sequential multi-class labeling in crowdsourcing,''
  \emph{IEEE Transactions on Knowledge and Data Engineering}, vol.~31, no.~11,
  pp. 2190--2199, 2018.

\bibitem{Androulakis2001}
I.~P. Androulakis, ``Dynamic programming: Stochastic shortest path problems.''
  2009.

\bibitem{settles-meeder-2016-trainable}
B.~Settles and B.~Meeder, ``A trainable spaced repetition model for language
  learning,'' in \emph{Proceedings of the 54th annual meeting of the
  association for computational linguistics (volume 1: long papers)}, 2016, pp.
  1848--1858.

\bibitem{9130935}
J.~Ke \emph{et~al.}, ``Learning to delay in ride-sourcing systems: a
  multi-agent deep reinforcement learning framework,'' \emph{IEEE Transactions
  on Knowledge and Data Engineering}, vol.~34, no.~5, pp. 2280--2292, 2020.

\bibitem{9031418}
Y.~Zhang \emph{et~al.}, ``Cost-sensitive portfolio selection via deep
  reinforcement learning,'' \emph{IEEE Transactions on Knowledge and Data
  Engineering}, vol.~34, no.~1, pp. 236--248, 2020.

\bibitem{upadhyay2018deep}
U.~Upadhyay, A.~De, and M.~Gomez~Rodriguez, ``Deep reinforcement learning of
  marked temporal point processes,'' \emph{Advances in Neural Information
  Processing Systems}, vol.~31, 2018.

\bibitem{Anderson1996ACT}
J.~R. Anderson, ``Act: A simple theory of complex cognition.'' \emph{American
  psychologist}, vol.~51, no.~4, p. 355, 1996.

\bibitem{NIPS2009_6bc24fc1}
H.~Pashler \emph{et~al.}, ``Predicting the optimal spacing of study: A
  multiscale context model of memory,'' \emph{Advances in neural information
  processing systems}, vol.~22, 2009.

\bibitem{10059206}
J.~Su \emph{et~al.}, ``Optimizing spaced repetition schedule by capturing the
  dynamics of memory,'' \emph{IEEE Transactions on Knowledge and Data
  Engineering}, 2023.

\bibitem{10.1145/3534678.3539081}
J.~Ye, J.~Su, and Y.~Cao, ``A stochastic shortest path algorithm for optimizing
  spaced repetition scheduling,'' in \emph{Proceedings of the 28th ACM SIGKDD
  Conference on Knowledge Discovery and Data Mining}, 2022, pp. 4381--4390.

\bibitem{Maddox2011TheRO}
G.~B. Maddox \emph{et~al.}, ``The role of forgetting rate in producing a
  benefit of expanded over equal spaced retrieval in young and older adults.''
  \emph{Psychology and aging}, vol.~26, no.~3, p. 661, 2011.

\bibitem{DBLP}
S.~Reddy \emph{et~al.}, ``Unbounded human learning: Optimal scheduling for
  spaced repetition,'' in \emph{Proceedings of the 22nd ACM SIGKDD
  international conference on knowledge discovery and data mining}, 2016, pp.
  1815--1824.

\bibitem{DBLP:journals/corr/abs-2102-04174}
A.~Nioche \emph{et~al.}, ``Improving artificial teachers by considering how
  people learn and forget,'' in \emph{26th International Conference on
  Intelligent User Interfaces}, 2021, pp. 445--453.

\bibitem{2021Reinforced}
V.~Vapnik and R.~Izmailov, ``Reinforced svm method and memorization
  mechanisms,'' \emph{Pattern Recognition}, vol. 119, p. 108018, 2021.

\bibitem{wang2022generalizationmemorization}
Z.~Wang and Y.-H. Shao, ``Generalization-memorization machines,'' \emph{arXiv
  preprint arXiv:2207.03976}, 2022.

\bibitem{Smola1998LearningWK}
A.~J. Smola and B.~Sch{\"o}lkopf, \emph{Learning with kernels}.\hskip 1em plus
  0.5em minus 0.4em\relax Citeseer, 1998, vol.~4.

\bibitem{788640}
V.~N. Vapnik, ``An overview of statistical learning theory,'' \emph{IEEE
  transactions on neural networks}, vol.~10, no.~5, pp. 988--999, 1999.

\bibitem{Vapnik2006EstimationOD}
V.~Vapnik, \emph{Estimation of dependences based on empirical data}.\hskip 1em
  plus 0.5em minus 0.4em\relax Springer Science \& Business Media, 2006.

\bibitem{2000New}
B.~Sch{\"o}lkopf \emph{et~al.}, ``New support vector algorithms,'' \emph{Neural
  computation}, vol.~12, no.~5, pp. 1207--1245, 2000.

\bibitem{Jayadeva2007Twin}
Jayadeva, R.~Khemchandani, and S.~Chandra, ``Twin support vector machines for
  pattern classification,'' \emph{IEEE Transactions on Pattern Analysis and
  Machine Intelligence}, vol.~29, pp. 905--910, 2007.

\bibitem{2002Least}
J.~A. Suykens and J.~Vandewalle, ``Least squares support vector machine
  classifiers,'' \emph{Neural processing letters}, vol.~9, pp. 293--300, 1999.

\bibitem{zhang2017understanding}
C.~Zhang \emph{et~al.}, ``Understanding deep learning (still) requires
  rethinking generalization,'' \emph{Communications of the ACM}, vol.~64,
  no.~3, pp. 107--115, 2021.

\bibitem{Belkin_2019}
M.~Belkin \emph{et~al.}, ``Reconciling modern machine-learning practice and the
  classical bias--variance trade-off,'' \emph{Proceedings of the National
  Academy of Sciences}, vol. 116, no.~32, pp. 15\,849--15\,854, 2019.

\end{thebibliography}
\end{document}